\documentclass[sigplan,nonacm]{acmart}
\usepackage[normalem]{ulem}

\usepackage{tikz}

\usepackage{color}
\usepackage{listings}
\usepackage{hyperref}
\usepackage{amsmath,amsfonts}
\usepackage{mathtools}
\usepackage[ruled, vlined, linesnumbered]{algorithm2e}
\usepackage{circledsteps}
\usepackage{comment}
\usepackage{booktabs}
\usepackage{multirow}
\usepackage{graphicx}
\usepackage{makecell}
\usepackage{titlesec}
\usepackage{pifont}
\newcommand{\cmark}{\ding{51}}%
\newcommand{\xmark}{\ding{55}}%

\usepackage{wasysym}

\definecolor{dmlgreen}    {RGB}{51,  160,  44}
\definecolor{dmlblue}     {RGB}{31,  120, 180}
\definecolor{dmlred}      {RGB}{202,   0,  32}
\definecolor{brown}       {RGB}{139,  69,  19}

\definecolor{deepblue}{rgb}{0,0,0.5}
\definecolor{deepred}{rgb}{0.6,0,0}
\definecolor{deepgreen}{rgb}{0,0.5,0}
\definecolor{mauve}{rgb}{0.58,0,0.82}
\definecolor{light-gray}{gray}{0.96}
\definecolor{aliceblue}{rgb}{0.94, 0.97, 1.0}

\usepackage[utf8]{inputenc}
\DeclareFixedFont{\ttb}{T1}{txtt}{bx}{n}{8} 
\DeclareFixedFont{\ttm}{T1}{txtt}{m}{n}{8}  

\lstnewenvironment{env_python}[2]
{
\lstset{
  language=Python,
  backgroundcolor=\color{light-gray},
  otherkeywords={self},        
  emph={to,tile,reuse_at,reorder,unroll,split,vectorize,quantize},          
  emphstyle=\footnotesize\color{dmlred},  
  belowcaptionskip=0.7\baselineskip,
  aboveskip=-2mm,
  belowskip=5mm,
  showstringspaces=false,
  columns=flexible,
  escapechar=@,
  basicstyle={\linespread{0.99}\fontencoding{T1}\footnotesize\fontfamily{lmtt}\fontseries{m}\selectfont},
  numbers={left},
  xleftmargin={2em},%
  framesep=5pt, 
  framexleftmargin={1em},
  numberstyle=\tiny\color{gray},
  keywordstyle=\color{blue},
  commentstyle=\color{deepgreen},
  stringstyle=\color{mauve},
  breaklines=true,
  breakatwhitespace=true,
  tabsize=3,
  frame=tb, 
  caption= #1, 
  label= #2
}
  \vspace{\baselineskip}
}
{}

\newcommand{\lstbg}[3][0pt]{{\fboxsep#1\colorbox{#2}{\strut #3}}}
\lstdefinelanguage{diff}{
  backgroundcolor=\color{aliceblue},           
  emph={},          
  emphstyle=\small\color{dmlred},  
  belowcaptionskip=0.7\baselineskip,
  aboveskip=0mm,
  belowskip=3mm,
  showstringspaces=false,
  columns=flexible,
  basicstyle={\linespread{1.1}\fontencoding{T1}\scriptsize\fontfamily{lmtt}\fontseries{m}\selectfont},
  numbers={left},
  xleftmargin={2em},%
  breaklines=true,
  breakatwhitespace=true,
  tabsize=3,
  morecomment=[f][\lstbg{red!20}]-,
  morecomment=[f][\lstbg{green!20}]+,
  morecomment=[f][\textit]{@@},
}

\usepackage{xcolor}

\makeatletter
\let\old@lstKV@SwitchCases\lstKV@SwitchCases
\def\lstKV@SwitchCases#1#2#3{}
\makeatother
\usepackage{lstlinebgrd}
\makeatletter
\let\lstKV@SwitchCases\old@lstKV@SwitchCases

\lst@Key{numbers}{none}{%
    \def\lst@PlaceNumber{\lst@linebgrd}%
    \lstKV@SwitchCases{#1}%
    {none:\\%
     left:\def\lst@PlaceNumber{\llap{\normalfont
                \lst@numberstyle{\thelstnumber}\kern\lst@numbersep}\lst@linebgrd}\\%
     right:\def\lst@PlaceNumber{\rlap{\normalfont
                \kern\linewidth \kern\lst@numbersep
                \lst@numberstyle{\thelstnumber}}\lst@linebgrd}%
    }{\PackageError{Listings}{Numbers #1 unknown}\@ehc}}
\makeatother

\newcommand{\Name}{{Slapo}\xspace}
\newcommand{\fx}{\texttt{torch.fx}\xspace}

\newcommand{\showcomments}{yes}
\newcommand\fixme[1]{
    \ifthenelse{\equal{\showcomments}{yes}}{\textcolor{red}{[#1]}}{\ignorespaces}
}
\newcommand\hz[1]{
    \ifthenelse{\equal{\showcomments}{yes}}{\textcolor{red}{[hz: #1~]}}{\ignorespaces}
}
\newcommand\cody[1]{
    \ifthenelse{\equal{\showcomments}{yes}}{\textcolor{blue}{[cody: #1~]}}{\ignorespaces}
}
\newcommand\shuai[1]{
    \ifthenelse{\equal{\showcomments}{yes}}{\textcolor{green}{[shuai: #1~]}}{\ignorespaces}
}
\newcommand\yida[1]{
    \ifthenelse{\equal{\showcomments}{yes}}{\textcolor{purple}{[yida: #1~]}}{\ignorespaces}
}
\newcommand\zz[1]{
    \ifthenelse{\equal{\showcomments}{yes}}{\textcolor{violet}{[zhiru: #1]}}{\ignorespaces}
}
\newcommand{\newtext}[1]{\textcolor{black}{#1}}
\newcommand{\revise}[1]{\textcolor{black}{#1}}

\AtBeginDocument{%
  }

\setcopyright{acmlicensed}
\copyrightyear{2024}
\acmYear{2024}
\acmDOI{XXXXXXX.XXXXXXX}

\acmConference[ASPLOS'24]{International Conference on Architectural Support for Programming Languages and Operating Systems}{April 27--May 01,
  2024}{San Diego, CA}
\acmISBN{978-1-4503-XXXX-X/18/06}

\begin{document}

\title{\Name: A Schedule Language for Progressive Optimization of Large Deep Learning Model Training}

\author{Hongzheng Chen}
\authornote{Part of the work done while interning at Amazon.}
\affiliation{%
  \institution{Cornell University}
  \city{Ithaca}
  \state{New York}
  \country{USA}
}
\email{hzchen@cs.cornell.edu}

\author{Cody Hao Yu}
\authornote{Work done while at Amazon.}
\affiliation{%
  \institution{Boson AI, Inc}
  \city{Santa Clara}
  \state{California}
  \country{USA}
}
\email{cody@boson.ai}

\author{Shuai Zheng}
\authornotemark[2]
\affiliation{%
  \institution{Boson AI, Inc}
  \city{Santa Clara}
  \state{California}
  \country{USA}
}
\email{shuai@boson.ai}

\author{Zhen Zhang}
\affiliation{%
  \institution{Amazon Web Services}
  \city{Santa Clara}
  \state{California}
  \country{USA}
}
\email{zhzhn@amazon.com}

\author{Zhiru Zhang}
\affiliation{%
  \institution{Cornell University}
  \city{Ithaca}
  \state{New York}
  \country{USA}
}
\email{zhiruz@cornell.edu}

\author{Yida Wang}
\affiliation{%
  \institution{Amazon Web Services}
  \city{Santa Clara}
  \state{California}
  \country{USA}
}
\email{wangyida@amazon.com}

\renewcommand{\shortauthors}{Chen et al.}

\begin{abstract}
Recent years have seen an increase in the development of large deep learning (DL) models, which makes training efficiency crucial. 
Common practice is struggling with the trade-off between usability and performance.
On one hand, DL frameworks such as PyTorch use dynamic graphs to facilitate model developers at a price of sub-optimal model training performance.
On the other hand, practitioners propose various approaches to improving the training efficiency by sacrificing some of the flexibility, ranging from making the graph static for more thorough optimization (e.g., XLA) to customizing optimization towards large-scale distributed training (e.g., DeepSpeed and Megatron-LM).

In this paper, we aim to address the tension between usability and training efficiency through separation of concerns.
Inspired by DL compilers that decouple the platform-specific optimizations of a tensor-level operator from its arithmetic definition, this paper proposes a schedule language, \Name, to decouple model execution from definition.
Specifically, \Name works on a PyTorch model and uses a set of schedule primitives to convert the model for common model training optimizations such as high-performance kernels, effective 3D parallelism, and efficient activation checkpointing.
Compared to existing optimization solutions, \Name \emph{progressively} optimizes the model ``as-needed'' through high-level primitives, and thus preserving programmability and debuggability for users to a large extent.
Our evaluation results show that by scheduling the existing hand-crafted optimizations in a systematic way using \Name, we are able to improve training throughput by up to 2.92$\times$ on a single machine with 8 NVIDIA V100 GPUs, and by up to 1.41$\times$ on multiple machines with up to 64 GPUs, when compared to the out-of-the-box performance of DeepSpeed and Megatron-LM.

\end{abstract}

\maketitle

\section{Introduction}
\label{sec:intro}

The demand of large deep learning (DL) models is surging in recent years as they demonstrate dominating model accuracy on a range of tasks in natural language processing (NLP)~\cite{rishi2021foundation,devlin2018bert,brown2020language,chowdhery2022palm} and computer vision~\cite{zagoruyko2016wide,dosovitskiy2021vit,liu2022swinv2}.
These models are normally invented in user-friendly DL frameworks like PyTorch~\cite{paszke2019pytorch} with dynamic model graphs\footnote{Dynamic graph DL frameworks construct the model graph on the fly when executing its forward computation instead of constructing the graph ahead of time.}, which by design lacks sufficient optimization for high-performance execution.
This issue becomes more and more critical as the size of models grows exponentially and so does the time of training.

In order to reduce the model training time, developers propose various kinds of optimization.
The first type of optimization is implemented manually in different layers of model training, such as inserting high-performance kernels~\cite{dao2022flashattention,shoeybi2019megatron,apex,xFormers2022} for computationally intensive operators on specific devices (e.g., NVIDIA GPUs), employing data, tensor, and pipeline parallelism~\cite{rajbhandari2020zero,shoeybi2019megatron,narayanan2019pipedream}, as well as activation checkpointing~\cite{chen2016checkpoint,marisa2021dtr,jain2020checkmate}, to efficiently distribute the training across multiple devices.
However, manual optimization introduces the following two challenges.

\noindent\textbf{Challenge~1: Generality --} 
Incorporating the above optimizations requires making intrusive changes to the model implementation, which means that the optimization is not easy to generalize to other models.
A new model, even with minimal change from the old one, may not be able to directly reuse the old optimization.
In addition, the optimized model becomes platform-specific, requiring developers to maintain multiple implementations to serve all requirements (e.g., training on different platforms and deploying on non-GPU devices).

\noindent\textbf{Challenge~2: Ease of Tuning --}
In practice, an optimization scheme has a number of configurations to tune (e.g., pipeline stages, number of activation checkpoints) to get a combination that results in the best performance. 
Developers need to identify tunable configurations in the implementation and modify the model to expose them for effective tuning.
This process can be tedious and error-prone especially when the model definition is closely tied to optimizations.

In addition to manual optimization, the other set of optimization approaches converts the DL model into a number of \emph{static graphs} and leverages DL compilers to automatically apply optimizations.
For example, JAX~\cite{jax2018github} is a DL framework powered by a compiler XLA~\cite{xla}. 
JAX traces the entire model to obtain a whole graph statically, on top of which the compiler can perform aggressive optimizations such as operator fusion, expression simplification, and even 3D parallelism~\cite{zheng2022alpa}.
Similarly, the recent release PyTorch 2.0~\cite{pytorch20} provides a compiler interface to trace PyTorch dynamic graph executions and construct static graphs in \fx~\cite{reed2022fx} for optimizations. 
While automatic optimization requires minimal engineering effort from model developers and addresses some of the challenges mentioned above, it also introduces two new challenges.

\noindent\textbf{Challenge~3: Programmability --} 
Working on static model graphs is limited by the requirement that everything must be statically analyzable and deterministic.
Frameworks may impose constraints on the users to facilitate the conversion to static graphs.
For example, JAX programming model requires pure Python functions, no in-place updates, etc., so developers may need to rewrite the model to meet these constraints in order to make it runnable~\cite{jax2018github}.
For another example, PyTorch 2.0 cannot trace through the collective operators like \texttt{all\_reduce} which are essential for distributed training~\cite{pytorch20}.
Moreover, it is usually non-trivial for developers to control or configure the optimizations in fine granularity, such as disabling certain rules, or excluding certain operators from a compiler pass.


\noindent\textbf{Challenge~4: Debuggability --} To make model implementation easy to understand and maintain, model developers usually implement layer modules (e.g., convolutional, fully connected, and attention layers) as building blocks, and use them to compose a model \emph{hierarchically}.
However, to expand the scope of optimization and improve performance, DL compilers operating on a static model graph often flatten the hierarchy to create a single-level dataflow graph, and rewrite certain operators (e.g., decomposing the \texttt{batch\_norm} op into a number of smaller ones).
This prevents developers from understanding and troubleshooting performance or convergence issues, as the optimized model may bear little resemblance to the original model implementation.

\smallskip
To address the challenges mentioned above, we propose \emph{\Name}\footnote{
\url{https://github.com/awslabs/slapo}
}, a \underline{\textbf{S}}chedule \underline{\textbf{LA}}nguage for \underline{\textbf{P}}rogressive \underline{\textbf{O}}ptimization,
designed for DL frameworks with dynamic model graphs.
\Name has the following major features.

\noindent\textbf{Decouple model execution from its definition.} To address Challenge 1, we decouple model execution (named ``schedule'') from its definition.
As a result, model developers can maintain the same model implementation, and performance engineers can optimize a model- or platform-specific schedule in a separate place. 
This idea is inspired by well-known domain-specific compilers -- Halide~\cite{jonathan2013halide} and Apache TVM~\cite{chen2018tvm} -- which propose widely adopted schedule languages that decouple tensor operator scheduling from its arithmetic definition.

\noindent\textbf{Auto-tuning.} A separate schedule also enables massive auto-tuning opportunities. Similar to AutoTVM~\cite{chen2018learning}, \Name provides a programming interface that allows developers to specify a set of tuneable knobs to form an efficient tuning space. The tuning space can then be explored by \Name auto-tuner to realize the optimal configuration, which addresses Challenge 2. Along this direction, \Name can also enable auto-scheduling as Ansor~\cite{zheng2020ansor}, and this is our planned future work.

\noindent\textbf{Progressive optimization.} \Name incorporates a ``trace by need'' approach that only traces a desired module to be a static graph for compiler-based aggressive optimizations.
The traced part can be expanded or shrunk progressively as needed.
Developers explicitly call the scheduling primitives to realize this, addressing Challenge 3.

\noindent\textbf{Structure-preserving scheduling.} Model developers usually define building blocks (e.g., convolutional or attention layers), and then compose them together to form a complete model. Consequently, developers often leverage this structure to analyze and debug the model.
\Name preserves this hierarchy when constructing the schedule (see \S\ref{sec:sch} for details), so that developers can easily locate the module and apply scheduling.
Also, as the model structure is preserved and optimization can be progressively applied, it facilitates the users to debug any performance and convergence issue, and a verifier (\S\ref{sec:verifier}) is provided to further aid debugging, addressing Challenge 4.

In summary, we make the following contributions:
\begin{itemize}
    \setlength\itemsep{0.2em}
    \item We propose \Name, a schedule language that decouples model execution from definition, and preserves model structure hierarchy to enable progressive optimization.
    \item \newtext{We define a comprehensive set of schedule primitives for \Name to cover prevalent optimizations in distributed training, and provide an extensible infrastructure for users to easily incorporate their own optimizations.}
    \item We design and implement a lightweight auto-tuner for further reducing the efforts required to identify the optimal schedule configurations for training efficiency.
    \item We evaluate \Name by training popular deep learning models with billions of parameters and compare \Name with the state-of-the-art (SOTA) distributed training frameworks such as DeepSpeed~\cite{rasley2020deepspeed} and Megatron-LM~\cite{shoeybi2019megatron}. With minimal programming effort, \Name is capable of scheduling the existing hand-crafted optimizations to achieve up to 2.92$\times$ speedup on a single machine with 8 NVIDIA V100 GPUs, and up to 1.41$\times$ speedup on multiple machines with 64 V100 GPUs, when compared to the out-of-the-box best baselines.
\end{itemize}

\section{Background and Motivation}
\label{sec:background}
In this section, we first introduce common practices of improving a DL model training efficiency for dynamic graphs (e.g., PyTorch~\cite{paszke2019pytorch}), followed by an end-to-end motivating example to illustrate the challenges of this process.

\subsection{Efficient Model Training}
\label{sub:training}

\noindent\textbf{High-performance kernel libraries.} To achieve high efficiency in deep learning model training, it is straightforward to leverage efficient kernels specifically optimized for particular hardware platforms (e.g., NVIDIA GPUs, Google TPUs, and AWS Trainium).
For example, there are many libraries~\cite{dao2022flashattention,shoeybi2019megatron,apex,xFormers2022} that provide efficient CUDA kernel implementations.
These libraries encapsulate kernels as DL framework-compatible modules for developers to replace the native implementation in their models.
In the case where there are no existing CUDA implementations, developers may leverage compiler-based solutions, such as TorchScript~\cite{torchscript} and TorchInductor~\cite{torchinductor}, to generate a kernel.

\noindent\textbf{Activation checkpointing.} Apart from compute optimization techniques, memory footprint optimization is also essential for training large models.
A large portion of memory consumption in the training process is contributed by forward activation tensors that are stored for the later gradient calculation.
By checkpointing some activation tensors and re-computing the rest of them in backward propagation, we are able to significantly reduce memory footprint, and thus support a larger batch size and higher training throughput.
This approach is called activation checkpointing and is originally proposed by \cite{chen2016checkpoint}.
Furthermore, existing works~\cite{jain2020checkmate,marisa2021dtr,korthikanti2022reducing,cody2023raf} also demonstrate that by carefully selecting which activations to checkpoint, we are capable of better utilizing device memory and achieving an even better throughput.

\noindent\textbf{Parallelism in distributed training.} When the model is too large to fit in a single device, training it in parallel in a distributed environment is inevitable.
The parallelism techniques are usually classified into three types: data parallelism, tensor parallelism, and pipeline parallelism. Both tensor and pipeline parallelism belong to a larger class called model parallelism.
\emph{Data parallelism} partitions training data, so each device trains the replicated model with different data~\cite{li2020ptddp,abadi2016tensorflow,rajbhandari2020zero}, and aggregates their partial gradients for parameter updating.
Since data parallelism replicates an entire model on each device, it is insufficient when the model size (i.e., total memory consumption of its parameters) is too large to fit on a single GPU.
In this case, \emph{tensor parallelism}, takes another approach by partitioning the model parameter tensor onto multiple devices~\cite{shoeybi2019megatron}.
However, it requires developers to explicitly use collective communication operators to scatter tensors and aggregate partial results.
For example, Megatron-LM~\cite{shoeybi2019megatron} is a widely used PyTorch-based framework that provides manual parallelized Transformer models~\cite{vaswani2017transformer} and is adopted to train extremely large models~\cite{zhang2022opt}.
Finally, \emph{pipeline parallelism}~\cite{huang2019gpipe,narayanan2019pipedream} partitions models by layers and groups them into a series of stages. By putting each stage on a different device, we can overlap the computation of multiple data batches.
To ensure correctness and performance, pipeline parallelism needs a specialized runtime to schedule and synchronize data.
These techniques are not mutually exclusive and can be combined. Combining all of them is known as \emph{3D parallelism}~\cite{narayanan2021megatronv2}.

\subsection{A Motivating Example}
\label{sub:example}
\begin{figure}[!htbp]
\centering
\includegraphics[width=\linewidth]{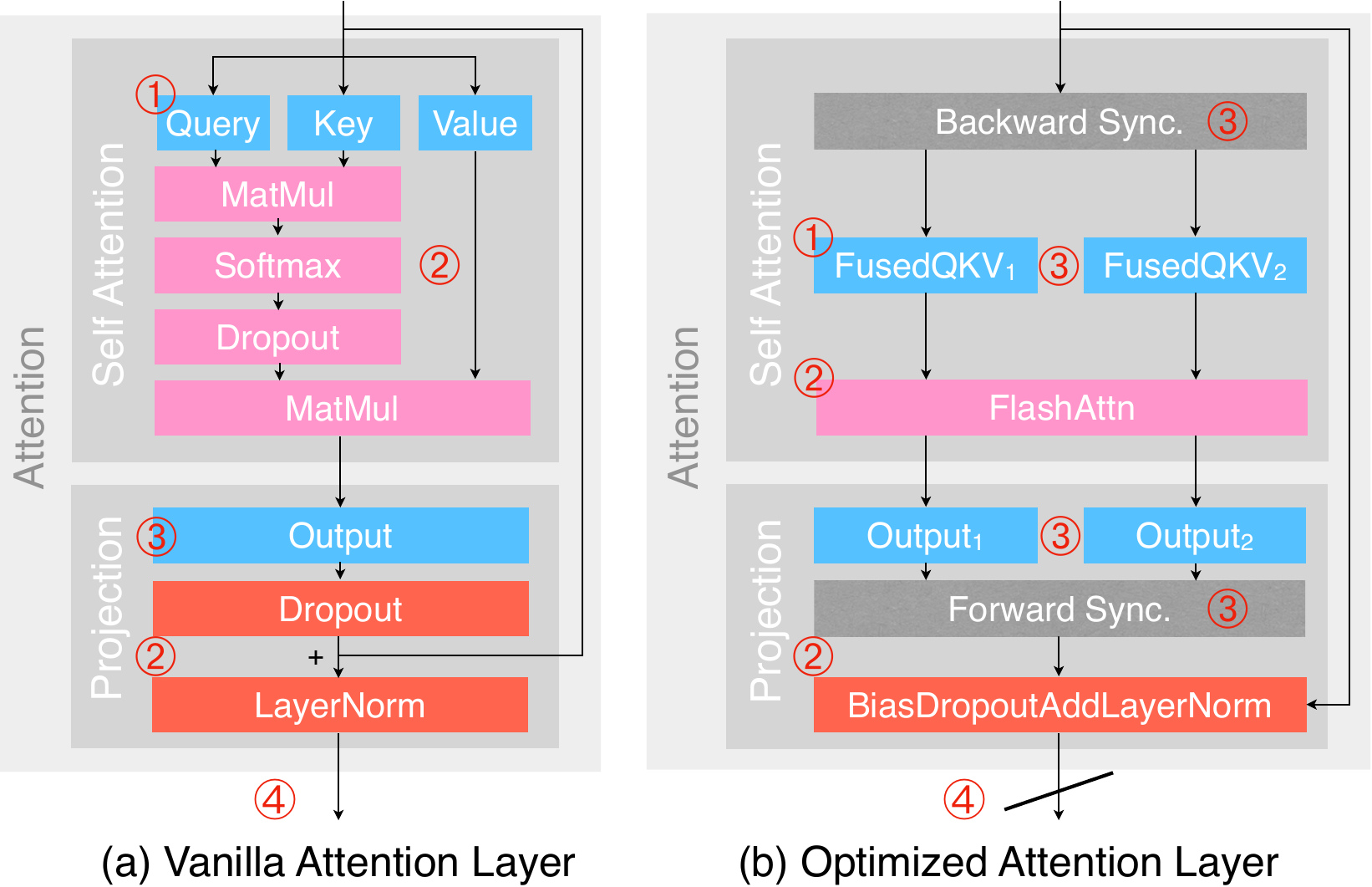}
\caption{An attention layer in BERT. The \texttt{Query}, \texttt{Key}, \texttt{Value}, and \texttt{Output} are \texttt{nn.Linear} modules containing learnable weights and biases. \Circled{1} - \Circled{4} indicate optimization points.}
\label{fig:attn}
\end{figure}

In this subsection, we use BERT~\cite{devlin2018bert} model implementation from HuggingFace Hub~\cite{wolf2019huggingface} to showcase how the above techniques are applied to a DL model for efficient training.
Fig.~\ref{fig:attn}(a) depicts the architecture of attention layer~\cite{vaswani2017transformer}, the core and most time-consuming module in BERT.
An attention layer is composed of two submodules -- \texttt{SelfAttention} and \texttt{Projection}.
We conduct a few steps to progressively improve the training efficiency of this attention layer and show the resulting implementation in Fig.~\ref{fig:attn}(b).

\begin{figure}[t]
\centering
\includegraphics[width=\linewidth]{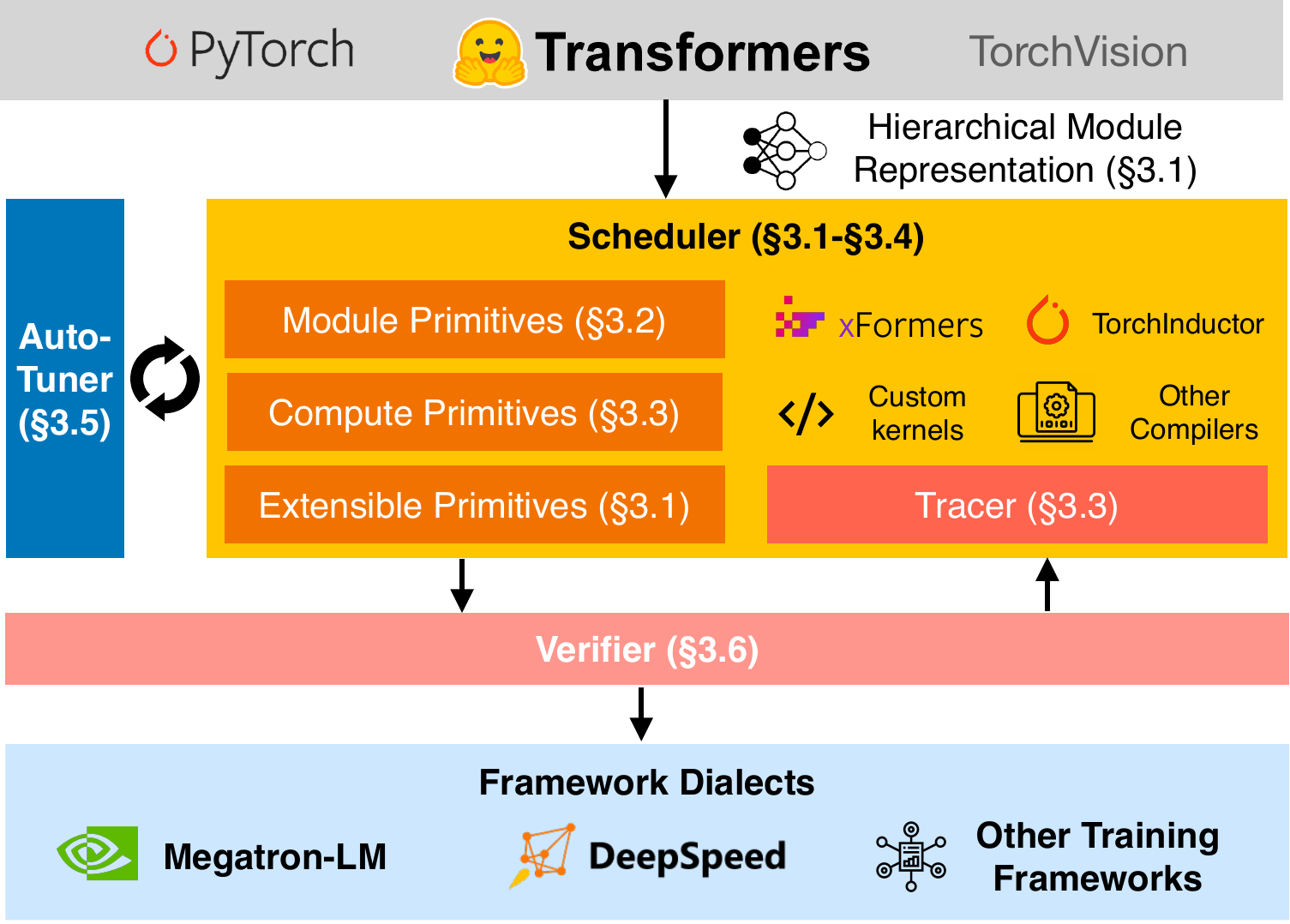}
\caption{Overview of \Name.}
\label{fig:overview}
\end{figure}

\noindent\textbf{\Circled{1} Fuse QKV.} By default, the \texttt{Query}, \texttt{Key}, \texttt{Value} in\\\texttt{SelfAttention} are three standalone \texttt{nn.Linear} modules, which may incur extra kernel launch overheads.
We replace them with a single \texttt{nn.Linear} module with their parameters concatenated, as shown in the following code snippet.

\smallskip
\begin{lstlisting}[language=diff]
def __init__(self, ...):
-   self.query = nn.Linear(n_hidden, n_head)
-   self.key = nn.Linear(n_hidden, n_head)
-   self.value = nn.Linear(n_hidden, n_head)
+   self.qkv = nn.Linear(n_hidden, n_head * 3)
def forward(self, hidden_states, ...):
-   query = transpose(self.query(hidden_states))
-   key = transpose(self.key(hidden_states))
-   value = transpose(self.value(hidden_states))
+   qkv = transpose(self.qkv(hidden_states))
+   query, key, value = split(qkv, 1, dim=-1)
\end{lstlisting}

\noindent\textbf{\Circled{2} Use efficient kernels.} The pink blocks in Fig.~\ref{fig:attn} are the core attention computation, which is also the bottleneck of performance and memory footprint.
A recent work flash attention~\cite{dao2022flashattention} proposes to compute the attention in a block-by-block fashion, so only a block of the intermediate attention tensor is generated at a time, significantly reducing the peak memory consumption, and thus can improve the training efficiency by enlarging the batch size.
The following code snippet shows how we replace the existing attention with the flash attention implementation provided by xFormers~\cite{xFormers2022}. The transpose and reshape operations are simplified.

\smallskip
\begin{lstlisting}[language=diff]
def forward(self, hidden_states, ...):
-   attn = query @ key.T
-   attn = attn / attn.shape[-1] ** 0.5
-   attn = dropout(softmax(attn), p)
-   attn = attn @ value
+   attn = xformers.ops.mem_eff_attention(...)
\end{lstlisting}

\noindent Note that this flash attention implementation only supports the latest NVIDIA GPUs with Volta, Ampere, and Hopper architectures, so once this kernel is used, the model is no longer compatible with other platforms.

Another optimization opportunity is the pattern in \texttt{Projection}. By default, the bias addition operation is contained in the \texttt{Output} module.
A more aggressive and more efficient way is to use a DL compiler (e.g., TorchScript~\cite{torchscript}, TorchInductor~\cite{torchinductor}) to fuse the pattern \texttt{BiasAdd-Dropout-ResidualAdd-}\\\texttt{LayerNorm} into a single kernel, as suggested by \cite{vaidya2022nvfuserexample}.

\noindent\textbf{\Circled{3} Tensor parallelism.} We then partition the \texttt{FusedQKV} and \texttt{Output} parameters onto multiple devices. 
Given the input of the attention module $X$, the weights of \texttt{FusedQKV} ($A$) and \texttt{Output} ($B$), we have self-attention function $f$:
\begin{equation*}
f(XA)B
=f\left(X\begin{bmatrix}A_1 & A_2\end{bmatrix}\right)
\begin{bmatrix}B_1 \\ B_2\end{bmatrix}
=f(XA_1)B_1 + f(XA_2)B_2\,.
\end{equation*}
Accordingly, we follow the convention of Megatron-LM \cite{shoeybi2019megatron} to shard the weights of \texttt{FusedQKV} in columns and shard the weights of \texttt{Output} in rows. We illustrate the latter in the following code snippet.

\smallskip
\begin{lstlisting}[language=diff]
def __init__(self, ...):
-   self.output = nn.Linear(n_hidden, n_hidden)
+   new_size = n_hidden // world_size
+   self.output = nn.Linear(new_size, n_hidden)
def forward(self, hidden_states):
    out = self.output(hidden_states)
+   dist.allreduce(out)
\end{lstlisting}

Since the output tensor only holds partial results after sharding, we need to conduct \texttt{all\_reduce} to aggregate outputs from different devices.

\noindent\textbf{\Circled{4} Pipeline parallelism.} To pipeline a BERT model and execute it on a SOTA pipeline runtime, we have to further manually partition the model to a sequence of sub-models (each of them includes a series of attention layers) by re-writing the top module\footnote{The code example is omitted due to page limit.}.

The above process poses a generality issue. 
Although developers have spent efforts on identifying and optimizing the performance bottleneck of a model with semantics preserved, this effort is hard to be reused by another model.
Furthermore, the above improved model is no longer compatible with a single device, unless we add control logic to only enable model parallelism on multi-GPU environments or maintain a separate single-device version. 
It also creates a barrier for the model deployment after training, because a model implementation with custom efficient kernels and parallelism may not be recognized by inference compilers.

In essence, the above pain points are the result of tightly coupling model definition and training/platform-specific optimizations. 
This motivates us to propose a schedule language that decouples model execution (i.e., schedule) from definition and provides easy-to-use primitives for optimizing large model training.
In fact, the idea of decoupling optimization has been widely accepted in DL compilers~\cite{jonathan2013halide,chen2018tvm,baghdadi2019tiramisu}, and opens opportunities for auto-tuning~\cite{chen2018tvm} and auto-scheduling~\cite{zheng2020ansor}.

\section{\Name Design}
\label{sec:name}

This section presents the design of \Name, our schedule language to progressively optimize DL model training using proposed primitives.
\Name decouples model definition from its training execution strategy for better portability.
\Name also abstracts out the common optimization techniques using a set of primitives to apply (or un-apply) one by one, lowering the bar for performance engineers to try out different optimization ideas.
Furthermore, \Name makes it possible to automate the performance tuning via hyperparameter search.

\begin{table}[t]
\centering
\caption{Comparison among \Name and other systems. DP, TP, and PP denote data, tensor, and pipeline parallelism respectively.
PT denotes PyTorch.
\newtext{Model coverage means how easily developers can leverage the programming system to optimize a new model.}}
\label{tab:comparison}
\resizebox{\columnwidth}{!}{%
\begin{tabular}{ccccccccc}\hline
& \multirow{2}{*}{\makecell[c]{Frame-\\work}} & \multirow{2}{*}{\makecell{Model\\Coverage}} & \multicolumn{3}{c}{3D parallelism} & \multirow{2}{*}{\makecell{Subgraph\\Opt.}} & \multirow{2}{*}{\makecell{Act.\\Ckpt.}} & \multirow{2}{*}{\makecell{Extensible\\Opt.}}\\
& & & DP & TP & PP & & \\
& & \S~\ref{sec:sch} & \multicolumn{3}{c}{\S~\ref{sub:sch-mod-sharding}
\& \S~\ref{sub:sch-compute-pipeline}} & \S~\ref{sub:sch-compute-partial} & \S~\ref{sub:sch-compute-partial} & \S~\ref{sec:sch}\\\hline
Megatron-LM~\cite{shoeybi2019megatron} & PT & \Circle & \cmark & \cmark & \cmark & \Circle & \LEFTcircle & \xmark\\
DeepSpeed~\cite{rasley2020deepspeed} & PT & \CIRCLE & \cmark & \xmark & \cmark & \Circle & \LEFTcircle & \xmark\\
Alpa~\cite{zheng2022alpa} & JAX & \CIRCLE & \cmark & \cmark & \cmark & \LEFTcircle & \Circle & \xmark\\
\texttt{pt.compile}~\cite{pytorch20} & PT &\LEFTcircle & \xmark & \xmark & \xmark & \CIRCLE & \Circle & \xmark\\
\Name & PT &\CIRCLE & \cmark & \cmark & \cmark & \CIRCLE & \CIRCLE & \cmark\\\hline
\end{tabular}%
}
\end{table}

Fig.~\ref{fig:overview} illustrates the overview of \Name.
\Name accepts a deep learning model implementation in a DL framework with dynamic graphs (e.g., PyTorch~\cite{paszke2019pytorch}) and parses the original model execution as its default schedule.
Then, developers make use of the schedule primitives for optimizations on top of the default schedule.
\newtext{\Name provides a comprehensive set of primitives that cover the prevalent distributed training optimizations, and Table~\ref{tab:comparison} compares \Name with other frameworks.}

We define the schedule language in \S\ref{sec:sch}, and present the scheduling in various scenarios in \S\ref{sec:sch-mod} and \S\ref{sec:sch-compute}.
The scheduling strategy can be auto-tuned to search for a configuration that achieves the best performance (\S\ref{sec:autotune}).
Meanwhile, \Name adopts a verifier (\S\ref{sec:verifier}) to ensure the functional correctness of all schedules.
After the schedule is determined and applied, the scheduled model can be trained on the runtime of the original DL framework (e.g., PyTorch), or if needed, on the dedicated runtime of existing distributed systems such as DeepSpeed~\cite{rasley2020deepspeed} pipeline.

\subsection{Schedule Language}
\label{sec:sch}

\begin{figure}[t]
    \centering
    \includegraphics[width=\linewidth]{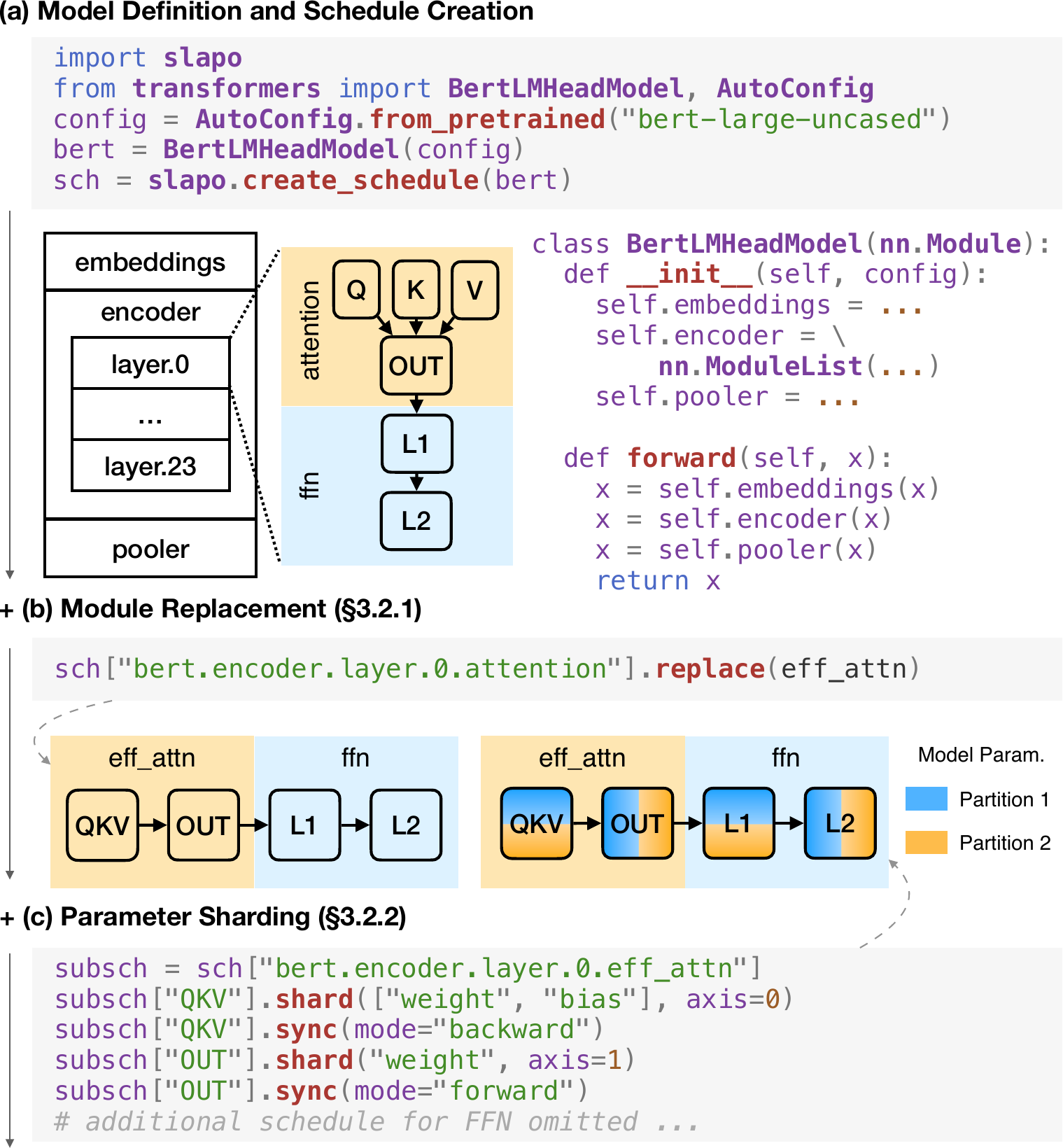}
    \caption{An example of scheduling modules and parameters of a BERT model. \texttt{ffn} is the Feed-Forward Network. \texttt{eff\_attn} refers to the replaced attention module. The weight matrix has a shape of \texttt{(output\_dim, input\_dim)}. Thus, sharding the weight matrix at \texttt{axis=0} is equivalent to partitioning the output dimension.}
    \label{fig:schedule_mod}
\end{figure}

Motivated by \S\ref{sub:example}, our goal is to let developers optimize models in a few lines of code without changing the model definition itself.
Fig.~\ref{fig:schedule_mod} presents the \Name schedule language with the BERT model from HuggingFace Hub~\cite{wolf2019huggingface} as an example.
As shown in Fig.~\ref{fig:schedule_mod}(a), most DL model developers define models with a hierarchical structure for better readability and easy maintenance.
The \texttt{\_\_init\_\_} constructor defines the configurations, submodules, and learnable parameters, and the \texttt{forward} method defines the forward computation (the backward computation is generated by the framework with automatic differentiation~\cite{paszke2019pytorch}).
Developers can then pass the created model into \Name and create a default \emph{schedule} that specifies \emph{how} to execute the model in the framework. 
The schedule preserves the hierarchical structure, and \texttt{create\_schedule} is applied recursively to all the submodules, so that developers can easily apply schedule primitives at arbitrary levels.

\begin{table}[t]
\centering
\caption{A summary of \Name built-in schedule primitives.}
\label{tab:primitives}
\resizebox{\columnwidth}{!}{%
\begin{tabular}{l|l}
\textbf{Primitives with Dynamic Graphs} & \textbf{Primitives with Static Graphs} \\ \hline\hline
\texttt{.replace(new\_mod)}             & \texttt{.replace(new\_mod, subgraph)}  \\ \hline
\texttt{.shard(param\_name, axis)}      & \texttt{.fuse(subgraph, compiler)}     \\ \hline
\texttt{.sync(type)}                    & \texttt{.pipeline\_split()}            \\ \hline
\texttt{.checkpoint()}                  & \texttt{.checkpoint(subgraph)}        
\end{tabular}%
}
\end{table}

\noindent\textbf{Built-In Primitives.}
As shown in Table~\ref{tab:primitives}, we categorize schedule primitives in two sets: whether or not they require a static graph to be generated.
On one hand, when scheduling at the module level, such as enabling activation checkpointing, replacing with an efficient alternative, and sharding learnable parameters, we do not change the computation specified in the \texttt{forward} method.
As a result, the schedule primitives in the left column of Table~\ref{tab:primitives} do not require a static graph, thus maximally avoiding the limitation of tracers.
We present the details of this scheduling in \S\ref{sec:sch-mod}.
On the other hand, scheduling the computation, such as operator fusion and pipeline parallelism, has to manipulate the \texttt{forward} method.
Thus, the schedule primitives in the right column of Table~\ref{tab:primitives} require the computation to be in a static graph, so we have to use \texttt{.trace()} prior to applying these primitives, as presented in \S\ref{sec:sch-compute}.
\revise{These primitives have covered existing optimizations ranging from parallelism schemes and compiler optimizations, which are general enough to support efficient training of different models, as demonstrated in \S\ref{sec:exp}.}

\noindent\newtext{\textbf{Extensible Primitives.}
In addition to the predefined primitives, users have the flexibility to incorporate their custom training optimization as a schedule primitive in \Name.
This can be achieved by inheriting the provided base primitive class as shown below.
During program execution, \Name dynamically registers the user-defined primitives, enabling seamless collaboration with other built-in primitives, the verifier, and the auto-tuner.}



\smallskip
\begin{lstlisting}[language=Python,numbers=left]
@slapo.register_primitive()
class UserDefinedPrimitive(slapo.Primitive):
  def __init__(self, name):
    ... # Initialize data structure and preconditions
  def apply(self, sch, **kwargs):
    ... # Transformation on the schedule
\end{lstlisting}




\subsection{Schedule Modules and Parameters}
\label{sec:sch-mod}

We first present scheduling a module and its parameters, which typically does not change the computation and thus does not require static graphs.

\subsubsection{Schedule Modules}
\label{sub:sch-mod}

For important workloads such as \texttt{attention} in Fig.~\ref{fig:schedule_mod}(a), researchers or hardware vendors may manually implement efficient kernels~\cite{dao2022flashattention,kao2023flat}. These highly customized, hand-crafted kernels sometimes could outperform the ones generated by DL compilers.
With \Name, we can use \texttt{.replace(new\_module)} primitive to replace a native implementation with an efficient one, where \texttt{new\_module} is the custom module to be replaced, as shown in Fig.~\ref{fig:schedule_mod}(b).

\newtext{Additionally, activation checkpointing is another important feature for large model training, as mentioned in \S\ref{sub:training}.
Existing frameworks~\cite{shoeybi2019megatron,rasley2020deepspeed} implement fixed strategies of checkpointing in their model definition and instantiate each layer with the same configuration, thus making it difficult to incorporate other checkpointing techniques~\cite{jain2020checkmate,korthikanti2022reducing}.
\Name decouples the checkpointing logic and offers a \texttt{.checkpoint()} primitive that can explicitly control whether a module should be checkpointed.
Consequently, \Name enables developers to flexibly adjust the number of checkpoints via our schedule primitive or leverage the auto-tuner for better memory and throughput trade-offs.}


\subsubsection{Parameter Sharding}
\label{sub:sch-mod-sharding}

\newtext{In \Circled{3} of \S\ref{sub:example}, we introduced the steps to enable tensor parallelism, which involves sharding parameters and aggregating outputs. This process is commonly known as the main challenge in adapting models for distributed execution.
The manual management of partitioning and communication within the model leads to a non-executable partitioned model when the number of devices changes, as well as makes synchronization with upstream model changes difficult.
While Megatron-LM~\cite{shoeybi2019megatron} provides tensor parallel modules for users, they are limited to specific models. If user-defined module operators differ from predefined modules, tensor parallelism cannot be utilized for distributed training.
}

\newtext{In contrast, \Name overcomes these limitations by enabling users to shard a parameter using the \texttt{.shard(param, axis)} primitive and aggregate results using the \texttt{.sync(type)} primitive.
The \texttt{type} can be ``forward'' (aggregate the forward activations) or ``backward'' (aggregate the gradients).
\revise{Notice \Name can efficiently capture the parameter and axes information covering the entire space of model partition, including 3D parallelism~\cite{shoeybi2019megatron,narayanan2021megatronv2} and other complicated parallelism schemes that an automatic compiler~\cite{zheng2022alpa} supports.
These primitives can be applied to arbitrary models and parameters, effectively addressing generality issues.}
It also does not require the model to be traceable since sharding does not modify the computation graph.
Fig.~\ref{fig:schedule_mod}(c) shows that implementing a complex tensor parallel program only requires five lines of schedule code without modifying the model definition.
\Name automatically shards parameters for different distributed environments and inserts synchronization points based on users' annotations.}
Meanwhile, we employ a verifier (\S\ref{sec:verifier}) to check correctness after scheduling.
In the future, we plan to develop an auto-scheduler that automatically generates these primitives.

\subsection{Schedule Computations}
\label{sec:sch-compute}

\begin{figure}[t]
    \centering
    \includegraphics[width=\linewidth]{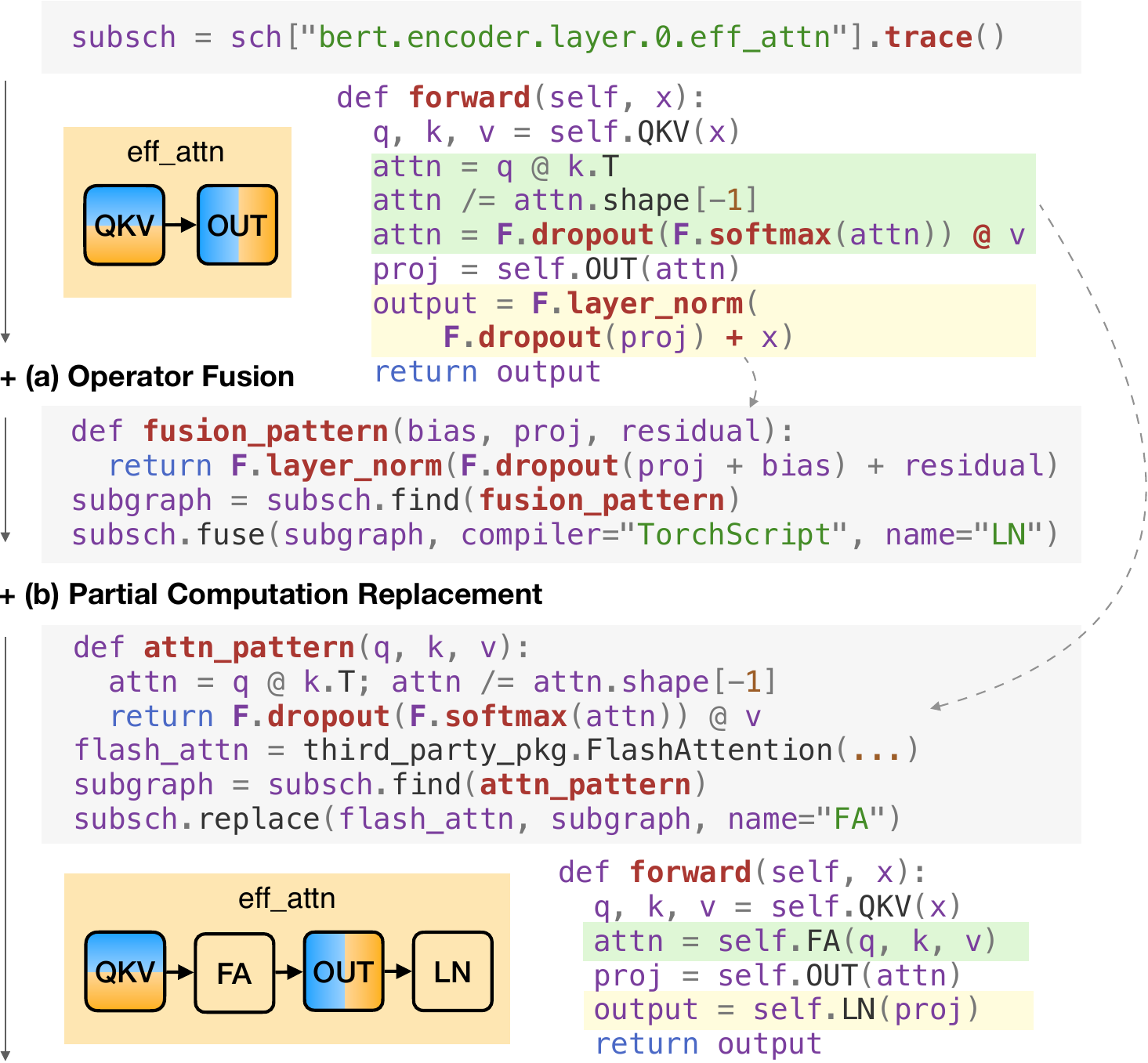}
    \caption{An example of scheduling computation of a BERT model. For illustration proposes, we do not show the entire IR of the traced \texttt{eff\_attn} module, but depict it in an identical \texttt{forward} function.
    \texttt{@} denotes dot product.
    The bias of the new \texttt{OUT} module is set as \texttt{None}, which has been fused into the \texttt{LN} module.}
    \label{fig:schedule_compute}
\end{figure}

The prerequisite of scheduling computations is tracing the \texttt{forward} method of the target module, and constructing a static graph in a certain intermediate representation (IR).
There are several approaches to obtaining the static graph IR. First, \textit{run-with-dummy-data}~\cite{torchscript} is an approach that directly executes the method with dummy inputs and captures all executed operators in order.
Second, \textit{AST-analysis}~\cite{torchscript} directly analyzes Python abstract-syntax-tree (AST) to obtain the static graph.
Third, \textit{just-in-time (JIT)}~\cite{suhan2021lazytensor,jax2018github} captures the execution graph in every training iteration, compiles it once, and reuses it in the rest process.
Finally, \textit{bytecode-analysis}~\cite{torchdynamo} hooks into the Python frame evaluation to construct the graph from Python bytecode.

In \Name, we define \texttt{.trace(leaves, flatten)} primitive on top of all approaches. This primitive lets developers configure the granularity and the form of a traced graph. Specifically, \texttt{leaves} indicate the submodules we will not trace into, and \texttt{flatten} indicates whether to flatten a traced static graph.
By default, the predefined modules (e.g., \texttt{nn.Linear}) in DL frameworks are all leaves, and we create the static graph in a hierarchical way.
The specification is then passed to the underlying tracing engine, and the traced module and submodules become static graphs so that compiler-related primitives can be enabled.
We show a traced BERT attention module in Fig.~\ref{fig:schedule_compute}.
In the rest of this subsection, we discuss the scheduling with static graphs.

\subsubsection{Partial Computation Scheduling}
\label{sub:sch-compute-partial}

The computation latency of a module is usually dominated by a part of its computational logic, such as \texttt{attn} and \texttt{output} in Fig.~\ref{fig:schedule_compute} (highlighted in yellow and green).
As a result, it is effective if developers can offload the performance bottleneck logic to efficient kernels or DL compilers.
Since most DL models usually have repetitive layers~\cite{devlin2018bert,zagoruyko2016wide}, \Name offers a \texttt{.find(regex\_or\_pattern\_fn)} primitive that performs pattern matching algorithm based on subgraph isomorphism~\cite{eppstein1995isomorphism}, with user-specified regular expression or a function with an identical subgraph.
This helps find all target subgraphs at once.
These subgraphs then can be scheduled in the same fashion with operator fusion, partial computation replacement, activation checkpointing, etc.


\noindent\textbf{Operator Fusion.} Operator fusion is an important optimization technique, as it can reduce the data transfer and kernel invocation overheads to improve the latency, throughput, and memory footprint.
\newtext{While existing DL compilers~\cite{xla,chen2018tvm,baghdadi2019tiramisu,torchscript} have well-established techniques for conducting operator fusion, distributed training frameworks~\cite{shoeybi2019megatron,rasley2020deepspeed} often cannot leverage them since they do not capture computation graphs and thus cannot apply automatic fusion mechanisms.
However, \Name can partially trace the module to avoid untraceable operators at the same time parallelizing the model, making it effortlessly compatible with existing fusion techniques.
As shown in Fig.~\ref{fig:schedule_compute}(a), \Name takes advantage of DL compilers by defining the \texttt{.fuse(subgraph, compiler)} primitive, where \texttt{compiler} indicates the DL compiler that will be used to generate a fused kernel of the subgraph.
We currently support a pattern-based fusion strategy and utilize TorchScript~\cite{torchscript} and TorchInductor~\cite{torchinductor} as DL compilers to enhance kernel performance.
}

\noindent\textbf{Partial Computation Replacement.} In addition to DL compilers, when the subgraph is the performance bottleneck in widely used models, developers may manually implement an efficient kernel and encapsulate it in a module. If this custom kernel achieves better performance than the one generated by a DL compiler, developers are capable of using \texttt{.replace(new\_mod, subgraph)} primitive to directly replace the corresponding computation logic with the custom one, such as Fig.~\ref{fig:schedule_compute}(b).
The decision of leveraging a DL compiler or a custom kernel can be made by developers with a one-line change. Developers can also rely on \Name auto-tuner (\S\ref{sec:autotune}) to realize a better one.




\noindent\textbf{Partial Activation Checkpointing.} In addition to enabling activation checkpointing for an entire module, as described in Section~\ref{sec:sch-mod}, \Name offers developers the flexibility to use the \texttt{.checkpoint(subgraph)} primitive, allowing checkpointing of specific subgraphs within the computation.
\newtext{This stands in contrast to existing PyTorch-based frameworks, which only support checkpointing at the module level due to their tightly coupled model implementation~\cite{rasley2020deepspeed,shoeybi2019megatron}.}
By providing this fine-grained control, we address the performance-memory trade-off dilemma, a topic extensively explored in various existing works~\cite{chen2016checkpoint,marisa2021dtr,jain2020checkmate}.

\subsubsection{Pipeline Partitioning}
\label{sub:sch-compute-pipeline}

In \S\ref{sec:sch-mod}, we demonstrated that \Name can achieve tensor parallelism using parameter sharding at the module level. However, it is not possible to achieve pipeline parallelism with the same approach, as it requires rewriting the top module, including its \texttt{forward} method, to be a sequence of submodules.

SOTA dynamic graph-based DL frameworks support pipeline parallelism in two steps. First, unlike the native runtimes of DL frameworks that use a single process to execute the model graph, pipeline parallelism requires launching one process per pipeline stage. Therefore, DL frameworks with pipeline parallelism support must provide their own runtime. Second, the model must be rewritten to follow a specific API convention. The rewritten implementation consists of a sequence of submodules, with outputs connecting to inputs between two consecutive submodules. This allows the DL framework to assign each module to a stage for execution. However, this process can be tedious for developers to prepare a model for pipeline parallelism~\cite{shoeybi2019megatron,rasley2020deepspeed}.

A recent work, PiPPy (Pipeline Parallelism for PyTorch)~\cite{pippy2022}, attempts to address this challenge by tracing the entire model to a static graph and partitioning the graph into a series of modules based on user annotations. However, this approach has limitations, as tracing the entire model can suffer from the limitations of the graph tracer, as discussed in \S\ref{sec:intro}. If a part of the model cannot be transformed into a static graph, the entire model cannot be partitioned. In contrast, \Name allows developers to configure the granularity and the form of the traced graph, meaning that developers can choose to only transform a few top-level modules into a static graph and use the \texttt{.pipeline\_split()} primitive to annotate the pipeline stage boundaries.

\begin{figure}[t]
    \centering
    \includegraphics[width=\linewidth]{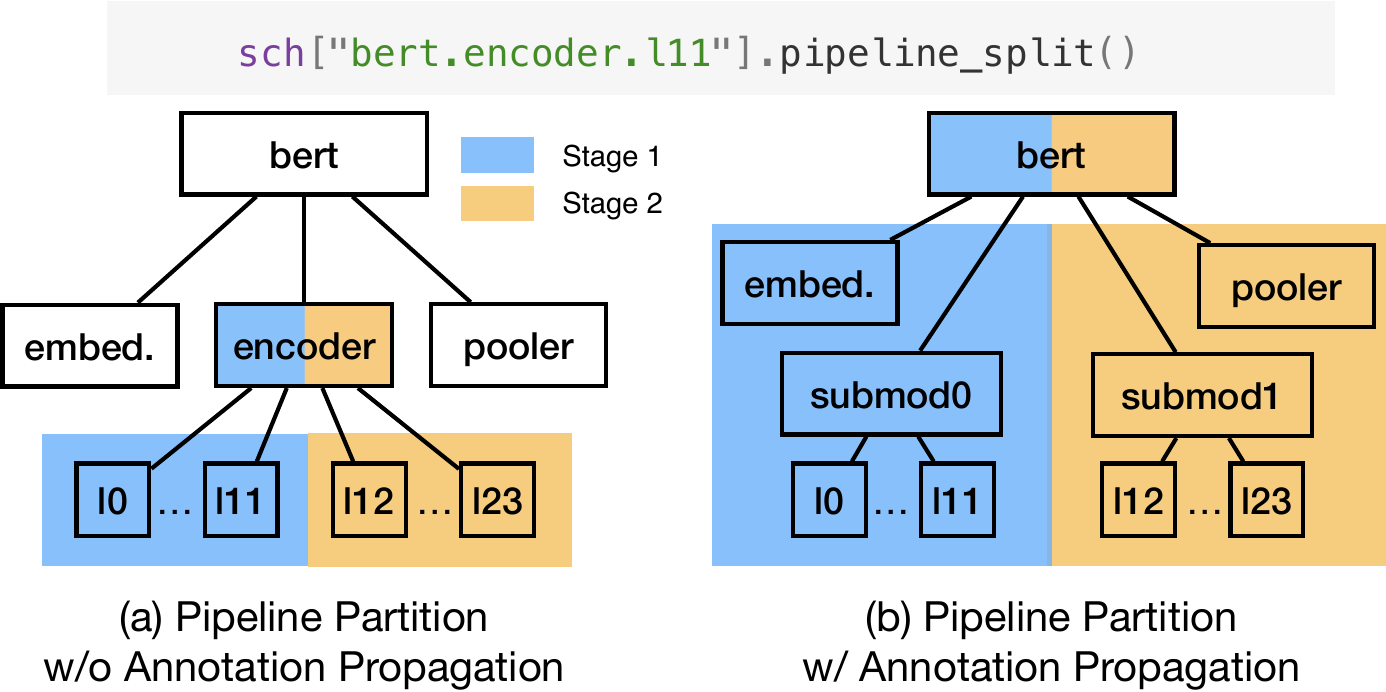}
    \caption{An example of partitioning a BERT model into two pipeline stages.}
    \label{fig:pipeline}
\end{figure}

We use the example in Fig.~\ref{fig:pipeline} for illustration.
To evenly split a BERT model with 24 attention layers in its encoder into two partitions, we can use \texttt{.pipeline\_split()} primitive\footnote{Where to insert pipeline splits to achieve optimal throughput is out of scope in this paper, but developers can use auto-tuner in \S\ref{sec:autotune} for exploration.} to annotate a stage boundary between layer 11 and 12 (0-based) in Fig.~\ref{fig:pipeline}(a).
In this case, only \texttt{encoder} module has to be traced, but not its submodules (e.g., \texttt{attention}) or siblings (e.g., \texttt{embeddings} and \texttt{pooler}).
We note that the untraceable, complex computation logic usually lies in core building block modules (e.g., \texttt{attention}), so limiting the tracing granularity makes our pipeline partitioning more applicable.

However, since \Name preserves the model structure hierarchy, it is non-trivial to partition a model into a sequence of modules with user-specified pipeline annotations.
Specifically, if we simply partition the model based on annotations as Fig.~\ref{fig:pipeline}(a), we fail to include \texttt{embeddings} and \texttt{pooler} modules in BERT architecture.
To generate the correct partitions, we propose an algorithm that propagates the pipeline annotations from the annotated submodule to the top module, so that all ancestor and descendant modules at all levels can be included. The algorithm is shown as follows.

\smallskip
\begin{lstlisting}[language=Python,numbers=left]
def partition(sch, common_parent_sch):
    seq_modules = partition_by_annotation(sch.module)
    parent_mod = sch.parent.module
    inline_and_annotate(seq_modules, parent_mod)
    if parent_mod != common_parent_sch.module:
        partition(parent_mod, common_parent_sch)

sub_schs = sch.get_all_sub_schedules_with_pipeline_split()
common_parent_sch = find_common_parent(sub_schs)
for sub_sch in sub_schs:
    partition(sub_sch, common_parent_sch)
partition(common_parent_sch, None)
\end{lstlisting}

We first retrieve all the subschedules whose children have pipeline partition annotation (L8). For the example in Fig.~\ref{fig:pipeline}, only a single subschedule \texttt{sch["bert.encoder"]} is returned. We then find the common parent module (L9, \texttt{bert} for Fig.~\ref{fig:pipeline}), and partition each submodule with the annotations (L10-11). After partitioning the current module (L2), we inline the partitioned module sequence and propagate the pipeline split annotations to its parent module (L4) so that the parent module now also has the annotations.
We perform this process recursively until the common parent module. At this point, all submodules have been partitioned and inlined, and the common parent module is not partitioned yet but is annotated by its children. Finally, we partition from the common parent module (L12) to the top module to finish the process as depicted in Fig.~\ref{fig:pipeline}(b).
\subsection{Auto-Tuning}
\label{sec:autotune}

Decoupling schedule from model definition enables auto-tuning. 
The combinations of schedule primitives provided by \Name can introduce a search space, which includes the decisions of the number of activation checkpoints and pipeline stages, whether to shard a parameter or replace a certain module/subgraph, etc.
Consequently, \Name provides an auto-tuner to explore the best combination given a particular training environment, further reducing the programming burden.

\begin{figure}[t]
    \centering
    \includegraphics[width=\linewidth]{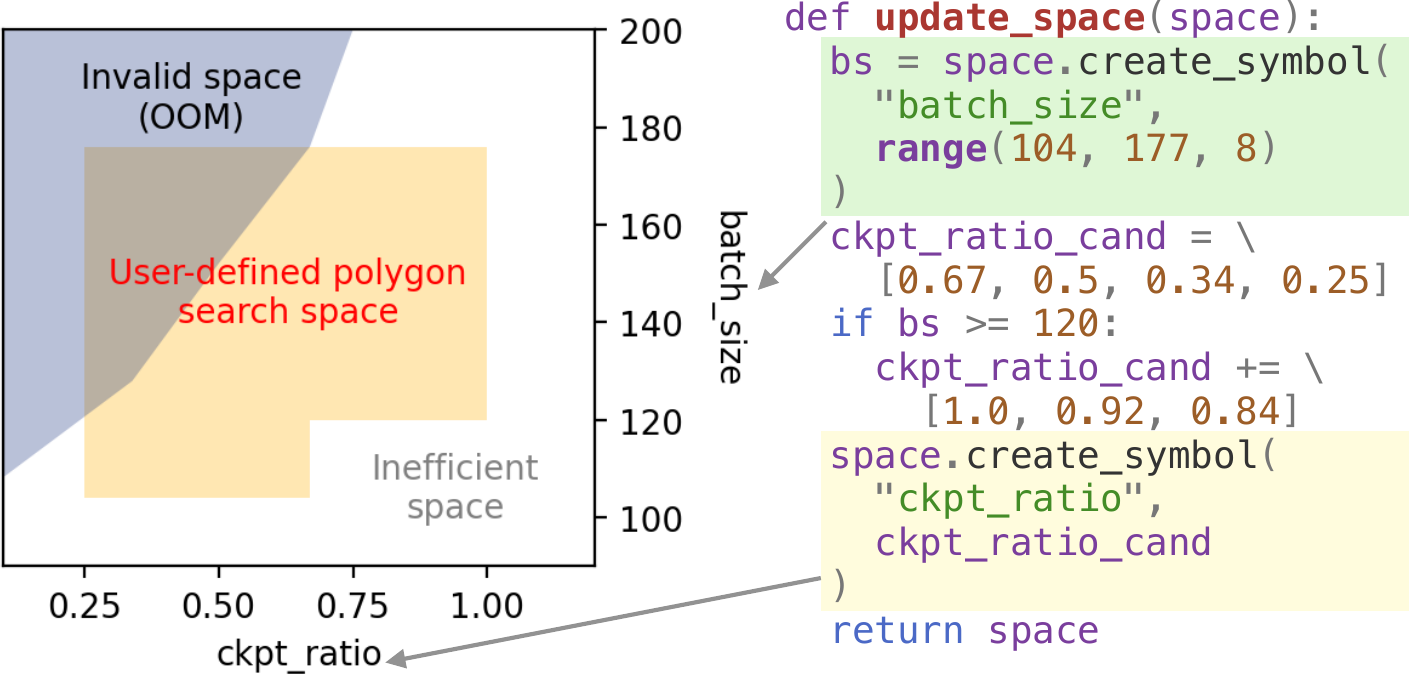}
    \caption{An example of a user-defined search space.}
    \label{fig:polygon}
\end{figure}

We provide symbolic variable constructions to help developers build a search space with arbitrary numbers of hyperparameters in a schedule.
Fig.~\ref{fig:polygon} depicts an example of a search space composed of batch size and the ratio of activation checkpointing.
In this example with two tunable parameters, we form the candidates of checkpoint ratio based on the batch size,
resulting in a polygon search space instead of a rectangle.
This allows developers to incorporate domain knowledge to prune inefficient configurations (the white region), and the invalid configurations (the gray region) can be quickly pruned by our auto-tuner.

With the search space constructed, the auto-tuner algorithm iteratively determines the values of all tunable parameters, schedules the model, and launches an evaluation script provided by developers to benchmark the performance and memory footprint.
Our auto-tuner leverages exhaustive search by default. Meanwhile, we also provide randomized coordinate descent~\cite{nesterov2012coordinate} for users to accelerate the tuning process.
We evaluate the efficiency of the search space and the tuning algorithm in \S\ref{sub:ablation} to show the effectiveness of our auto-tuner.


\subsection{Verification}
\label{sec:verifier}

To achieve high usability, \Name primitives are flexible to schedule modules and computations. However, it is possible for developers to schedule a model incorrectly. For example, the replaced module may require a different data layout but developers do not provide the corresponding layout transformation logic, or developers insert the output aggregation to an invalid point when sharding a parameter tensor.

To guarantee the scheduled model is executable and maintains its numerical correctness, \Name includes the following verification stages.
First, before applying the schedule, we validate the sequence of schedule primitives with a set of predefined rules in each primitive. 
For example, a \texttt{.sync()} primitive must have a corresponding \texttt{.shard()} primitive beforehand; primitives for distributed training can only be specified in a distributed environment with multiple devices; primitives that require static graphs must have a corresponding \texttt{.trace()} primitive in advance.
If any of the rules are violated, the verifier raises an error and stops the rest scheduling process.

\newtext{Second, we need to assure the numerical correctness of the scheduled model.
We provide a \texttt{.verify(*schs)} primitive for users to conduct differential testing~\cite{McKeeman1998DifferentialTesting} on different schedules.
The verifier generates random inputs and feeds in different scheduled models.
This validates both sharded parameter and tensor shapes, as well as confirming consistent outputs with the vanilla model.
It works in a distributed environment without altering the model and highlights problematic primitives for improved debuggability.
}

\section{Implementations}
\label{sec:impl}

We implemented \Name with $\sim$3K LoC in Python on top of PyTorch~\cite{paszke2019pytorch} to benefit from its dynamic graph and usability. In this section, we highlight some implementation details. 

\noindent{\textbf{Static Graph Tracing.}} Our tracer and the static graph are based on \fx~\cite{reed2022fx}, which captures PyTorch models via symbolic tracing and constructs a static graph with a 6-instruction Python-based intermediate representation (IR). Like other DL compilers, \fx tracer also has unsupported coding styles when capturing the graph, such as certain types of control flow and data structures. It also flattens the IR and discards all the model structural hierarchy.
Thus, instead of simply invoking \fx tracer from the top module, we invoke the tracer module by module and carefully maintain the hierarchy.
When a particular module cannot be traced by \fx, developers can simply disable the corresponding schedule primitives that require a static graph while the rest primitives can still be applied.
\newtext{Since we directly transform the modules and computation graphs and feed the scheduled model to the PyTorch runtime, \emph{no} additional overhead is introduced during execution.}


\noindent{\textbf{Framework Dialects.}} \Name scheduled model is by design compatible with native PyTorch runtime and can be executed on PyTorch directly.
To integrate with the dedicated runtime of SOTA distributed systems for certain parallelism (e.g., pipeline), we also implemented two framework dialects for Megatron-LM~\cite{shoeybi2019megatron} and DeepSpeed~\cite{rasley2020deepspeed}. 
\newtext{These systems require either wrapping the model with their custom module or adhering to specific input/output formats.
For instance, DeepSpeed~\cite{rasley2020deepspeed} pipeline runtime requires a single tuple of inputs and outputs in each pipeline stage, so our DeepSpeed dialect includes automatic logic to (1) unpack the inputs from the previous stage and encapsulate the outputs as a tuple for the next stage,
and (2) perform liveness analysis to bypass the tensors that are \emph{not} required by the current stage but are required by subsequent stages.
By leveraging the provided dialects, users only need to specify the target framework in \Name without modifying the model definition.}

\section{Evaluation}
\label{sec:exp}

In this section, we evaluate \Name on different training configurations, in terms of the number of GPUs, number of instances, and parallelism strategies, to demonstrate \Name is able to align or even outperform existing solutions while preserving usability.
\newtext{Note that \Name does \emph{not} change the semantics of models and optimizers, so model convergence is \emph{not} affected.
We also provide ablation studies to show the effectiveness of the schedule primitives and the auto-tuner.}



\noindent\textbf{Setups.}
All experiments are conducted on Amazon EC2 p3 instances. 
More specifically, we use p3dn.24xlarge instances with 8$\times$NVIDIA V100 32GB GPUs for single-node evaluations, and use at most 8$\times$p3dn.24xlarge instances for multi-node evaluations.
GPUs in these instances are connected via NVLink, which provides 300 GB/s theoretical aggregated GPU interconnect bandwidth, and the inter-node bandwidth is 100 Gbps.
The software environment includes CUDA 11.7, PyTorch v2.0.1, Megatron-LM (git-hash 0bb597b), DeepSpeed (v0.9.4), HuggingFace v4.28.1, and NCCL v2.14.3.

\noindent\textbf{Models and Metrics.}
\revise{We apply schedules to a set of popular PyTorch models from HuggingFace Hub~\cite{wolf2019huggingface} and torchvision~\cite{torchvision2016}, covering language models and image classification models to demonstrate the generality of \Name.
BERT and RoBERTa are encoder-only Transformer models.
GPT and OPT are decoder-only Transformer models.
T5 has both encoders and decoders.
WideResNet is a convolutional neural network.}
Detailed model settings are shown in Table~\ref{tab:models}.
Other models like graph neural networks~\cite{liu2023bgl,swapnil2021p3} require partitioning the graph structure which is out of our scope.
All models in the experiment are trained by AdamW optimizer~\cite{loshchilov2017decoupled} with mixed precision, and the micro-batch size (i.e., the number of samples per data parallel rank) is selected based on the memory footprint maximizing the system performance.
We use the training throughput (the number of total processed samples per second) as our evaluation metric.
For each setting, we train the models for tens of steps and take the average throughput after discarding the first few warm-up steps.

\begin{table}[!htbp]
\caption{Models used in the single-node experiments. \# of params shows the model size. MLM $=$ Mask language modeling. CLM $=$ Causal language modeling. Seq2Seq $=$ Sequence-to-Sequence modeling. IC $=$ Image Classification.}
\label{tab:models}
\centering
\resizebox{\columnwidth}{!}{
\begin{tabular}{ccccc}\Xhline{2\arrayrulewidth}
\textbf{Model} & \textbf{Task} & \begin{tabular}{c}\textbf{\# of params}\\\textbf{(Billion)}\end{tabular} & \begin{tabular}{c}\textbf{Seq Length} /\\\textbf{Image Size}\end{tabular} & \textbf{Precision}\\\hline
BERT~\cite{devlin2018bert} & MLM & 0.96 & 512 & FP16 \\\hline
RoBERTa~\cite{liu2019roberta} & MLM & 1.3 & 512 & FP16\\\hline
GPT~\cite{radford2019gpt2} & CLM & 2.86 & 1024 & FP16\\\hline
OPT~\cite{zhang2022opt} & CLM & 2.69 & 1024 & FP16\\\hline
T5~\cite{2020t5} & Seq2Seq & 2.85 & 1024, 512 & FP16\\\hline
WideResNet~\cite{zagoruyko2016wide} & IC & 2.4 & 3$\times$224$\times$224 & FP32\\
\Xhline{2\arrayrulewidth}
\end{tabular}
}
\end{table}

\begin{figure*}[t]
\centering
\begin{minipage}{0.75\textwidth}
\includegraphics[width=\linewidth]{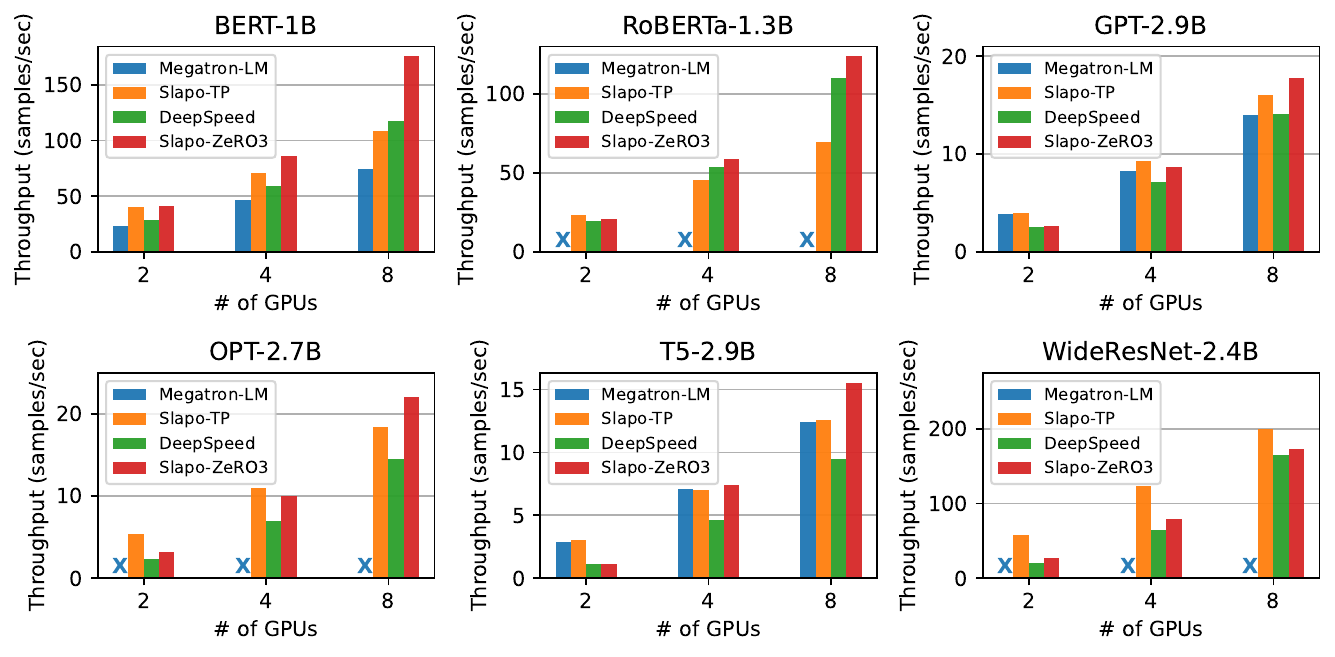}
\caption{Training throughput on Amazon EC2 p3dn.24xlarge instance with 8 V100 GPUs.\\``\Name-TP'' uses tensor parallelism to align Megatron-LM. ``\Name-ZeRO3'' uses DeepSpeed\\ZeRO-3 as the parallelism technique. ``X'' denotes not supported by the framework.}
\label{fig:8_v100}
\end{minipage}%
\begin{minipage}{0.25\textwidth}
\centering
\includegraphics[width=0.97\linewidth]{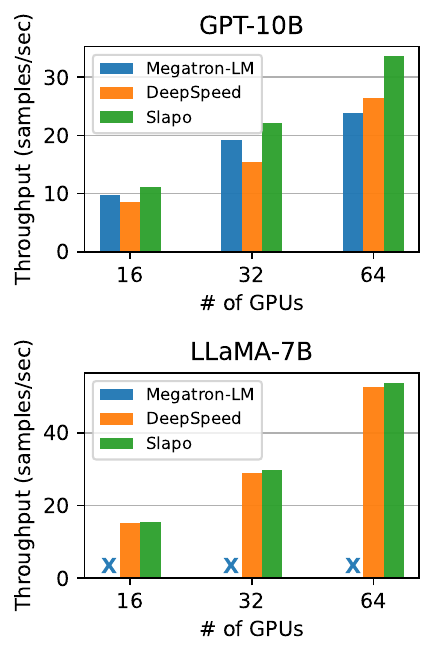}
\caption{Training throughput of different frameworks on up to 64 V100 GPUs.}
\label{fig:multinode}
\end{minipage}
\end{figure*}

\subsection{Evaluation on A Single Machine}
\label{sec:eval:multi-gpu}
This subsection evaluates the end-to-end training efficiency on 2, 4, and 8 NVIDIA V100 32GB GPUs in a single p3dn.24xlarge instance to showcase the effectiveness of \Name.

\noindent\textbf{Systems.}
We select Megatron-LM v2~\cite{narayanan2021megatronv2} as a strong baseline, which is a SOTA system built on top of PyTorch for training large Transformer-based language models on GPUs.
Megatron-LM implements its own data loader and optimizer for better training efficiency.
In addition, it implements popular Transformer models with tensor parallelism as well as efficient customized CUDA kernels.
We also choose DeepSpeed~\cite{rasley2020deepspeed} as another baseline. DeepSpeed is a SOTA framework that incorporates ZeRO-powered data parallelism (ZeRO-DP)~\cite{rajbhandari2020zero}, which applies to arbitrary PyTorch models and is widely used to train large models.
\revise{We tune both baseline systems by changing the configurations such as batch sizes and activation checkpoints to maximize their performance.}

\newtext{We focus our experiments on frameworks capable of training models that cannot be fit in a single device, and thus traditional data-parallel frameworks~\cite{disttf,jiang2020byteps,sergeev2018horovod} are not considered.
Additionally, JAX-based frameworks such as Alpa~\cite{zheng2022alpa} are excluded from the comparison since they are not directly comparable to PyTorch-based frameworks.
Alpa also does not exhibit performance advantages when compared to Megatron-LM for the tested models in Table~\ref{tab:models}, as the model architectures are regular~\cite{zheng2022alpa}.}
As \Name is agnostic to parallelism strategies, we evaluate two configurations for every model to show that \Name is compatible with the existing distributed training frameworks.
Specifically, ``\Name-ZeRO3'' schedules models with ZeRO-3~\cite{rajbhandari2020zero} that automatically partitions optimizer states, gradients, and parameters to enable memory-efficient data parallelism; while ``\Name-TP'' schedules them to enable tensor parallelism.
For each configuration, we auto-tune the checkpointing ratio along with the batch size.

\noindent\textbf{Results.}
We first compare two baselines, Megatron-LM and DeepSpeed ZeRO-3, in Fig.~\ref{fig:8_v100}.
It is worth noting that Megatron-LM officially only supports three (i.e., BERT, GPT, and T5) out of the six models listed in Table~\ref{tab:models}.
Comparing these three models, we find that 
no one solution is always superior to the other, highlighting the importance of \Name which enables developers to easily implement the best parallelism strategies using schedules for different models.

As shown in Fig.~\ref{fig:8_v100}, \Name can always perform the best and achieve up to 2.92$\times$ speedup compared to the baselines.
\Name-TP achieves throughput gains of 1.02$\times$ to 1.46$\times$ on 8 GPUs for the models supported by Megatron-LM.
Notably, for the BERT model, \Name-TP can achieve a speedup of up to 1.73$\times$.
We employ the tensor parallelism scheme proposed by Megatron-LM to shard both attention and MLP layers, thereby ensuring alignment of multi-device performance.
While Megatron-LM implements all the customized kernels within the framework, \revise{\Name captures subgraphs and enables additional optimization opportunities powered by deep learning compilers, thus leading to higher performance.}
We also note that \Name-TP does not significantly outperform Megatron-LM on GPT and T5 models.
The discrepancy can be attributed to variations in model implementations between Megatron-LM and HuggingFace, which include differences in position embedding, linear bias, layer norm, etc.


The model difference is eliminated when comparing \Name-ZeRO3 against DeepSpeed since both frameworks run the same HuggingFace models.
As illustrated in Fig.~\ref{fig:8_v100}, \Name-ZeRO3 consistently outperforms DeepSpeed by a margin of 1.04$\times$-1.64$\times$.
\revise{The speedup is primarily attributed to the utilization of efficient kernels, which are not included by default in the DeepSpeed training pipeline due to the need for extensive model and parameter rewriting.}
It is important to note that DeepSpeed supports only fully checkpointed layers implemented in HuggingFace, whereas \Name can conduct selective checkpointing. By combining this with the auto-tuner, we can consistently achieve higher performance compared to DeepSpeed.
\revise{We further conduct ablation study in \S\ref{sub:ablation} to demonstrate the performance improvements of different optimizations.}

\subsection{Evaluation on Multiple Machines}
This subsection presents the results of distributed training performance in multi-machine setups.

\noindent\textbf{Systems and Setups.} The testbed in this evaluation is\\p3dn.24xlarge instances with 8 NVIDIA V100 32 GB GPUs each.
We again use DeepSpeed and Megatron-LM v2, as baselines.
Following the common practices~\cite{rajbhandari2020zero, narayanan2021megatronv2}, we use ZeRO-3 for DeepSpeed and set model-parallel and pipeline-parallel size as 8 and 2, respectively, for Megatron-LM.
We consider the strong scaling efficiency, in which the global batch size is fixed. 
Typically, distributed training for large models runs on hundreds to thousands of GPUs and uses global batch sizes up to thousands out of consideration for model quality~\cite{narayanan2021megatronv2, brown2020language,chowdhery2022palm}.
We fix the global batch size at 256 for our clusters with up to 64 GPUs, and tune the micro-batch size of each system for the best performance.
The evaluation uses a GPT-10B model~\cite{radford2019gpt2} and a LLaMA-7B model~\cite{touvron2023llama}. The input sequence length for the experiments is 1024.




\noindent\textbf{Results.}
\newtext{First of all, Fig.~\ref{fig:multinode} reaffirms that no single parallelism strategy consistently performs best for different GPU configurations.
In contrast, \Name consistently outperforms the best available baselines by up to 1.41$\times$ when training the GPT-10B model.
This is due to \Name's flexibility, as it is not bound to a specific parallelism strategy or framework.
It allows us to choose the most effective strategy and incorporate additional optimizations.
By leveraging \Name, we can readily implement 3D parallelism for emerging models like LLaMA~\cite{touvron2023llama,touvron2023llama2}, which would require significant engineering efforts to be supported in Megatron-LM.
Nevertheless, \Name achieves a limited speedup over DeepSpeed in the case of LLaMA-7B.
According to our profiling, this is because the majority of ZeRO-3 overhead, weight all-gather, is moderate in the 7B-scale model when compared to 3D parallelism.
}
\subsection{Usability Study}
\label{sub:loc}

\begin{table}[t]
\centering
\caption{The total Line of Code required to implement high-performance schedules (sch) for models in Fig.~\ref{fig:8_v100} and Fig.~\ref{fig:multinode}.}
\label{tab:loc}
\resizebox{\columnwidth}{!}{%
\begin{tabular}{cc|cc|cc|cc}\Xhline{2\arrayrulewidth}
\textbf{Model} & \textbf{Sch} & \textbf{Model} & \textbf{Sch} & \textbf{Model} & \textbf{Sch} & \textbf{Model} & \textbf{Sch}\\\Xhline{2\arrayrulewidth}
BERT & 21 & GPT & 10 & T5 & 11 & LLaMA & 11\\
RoBERTa & 21 & OPT & 10 & WideResNet & 12\\
\Xhline{2\arrayrulewidth}
\end{tabular}%
}
\end{table}

\newtext{To demonstrate the improved usability of \Name, we present the line of code (LoC) count required to implement high-performance schedules in Table~\ref{tab:loc}.
Typical model implementations on the HuggingFace Hub consist of over 1,000 lines of code, making it impractical for developers to directly modify the internal modules scattered in different places to accommodate various hardware environments.
\Name offers a user-friendly interface that allows developers to incorporate the latest optimization techniques without altering the original model definition.
In most cases, users can achieve complicated distributed optimizations with about 20 lines of code, significantly reducing the coding burden.
Moreover, certain schedules can be shared among models with similar architectures, such as BERT and RoBERTa.
Even non-expert users can benefit from our predefined schedule templates to attain high performance.
For an illustrative example schedule on the BERT model, please refer to Supplemental Material~\textcolor{red}{A}.}

\revise{
To further illustrate how \Name can accommodate various optimizations through the extension of schedule primitives, we analyze its utilization by initial \Name users.
Our investigation assesses several pull requests submitted by different teams, each aiming at adding support for new primitives in our internal repository.
We selected several of them and requested detailed reports on the optimization scenarios facilitated by these new primitives, as well as the development efforts required to integrate them.
The results are shown in Table~\ref{tab:new_primitives}.
The proposed new primitives have been successfully applied to a diverse range of optimization tasks.
More importantly, these primitives can be swiftly implemented through the extensible primitive interface in \S\ref{sec:sch}, often requiring only a single day of development effort.
Even the development of an automatic build system for binding CUDA kernels can be accomplished within the same timeframe.
}

\begin{table}[htbp]
\caption{The development efforts of introducing new schedule primitives in \Name.
LoC is the line of code for implementing these optimizations as primitives.
The development time does \emph{not} include comprehensive testing on various models.}
\label{tab:new_primitives}
\resizebox{\columnwidth}{!}{%
\begin{tabular}{cccc}\Xhline{1pt}
\textbf{New primitives} & \textbf{Description} & \textbf{LoC} & \makecell[c]{\textbf{Approx.}\\\textbf{Develop Time}}\\\Xhline{1pt}
\texttt{.quantize()} & \makecell[c]{Replace a module with\\a predefined quantized module\\for quantization-aware training} & 11 & 1 hour\\\hline
\texttt{.bind()} &
\makecell[c]{Bind a module with\\
a CUDA kernel implementation} & 95 & 1 day\\\hline
\texttt{.cudagraphify()} &
\makecell[c]{Use CUDA graph~\cite{cudagraph} to\\
reduce kernel launch overheads} & 16 & 1 hour\\\Xhline{1pt}
\end{tabular}
}
\end{table}

\subsection{Ablation Study}
\label{sub:ablation}
We design an ablation study for applied optimizations to investigate the performance gain of the schedule.

\begin{figure}[t]
\centering
\includegraphics[width=0.9\linewidth]{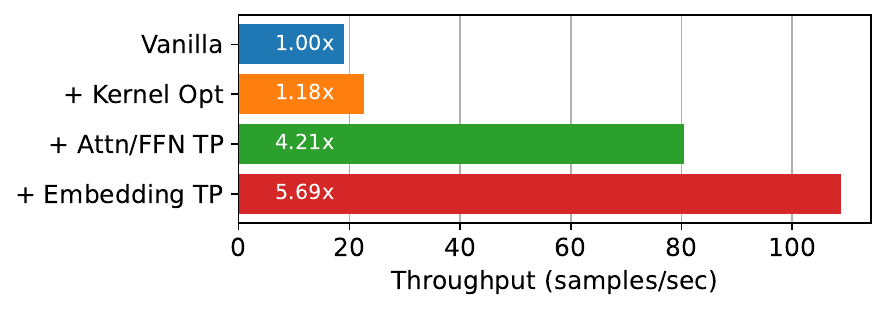}
\caption{Ablation study with HuggingFace BERT model.}
\label{fig:ablation}
\end{figure}

\noindent\textbf{Schedule Primitives.}
We start from a vanilla HuggingFace BERT model that is only capable of running on a single device.
\newtext{As evident in Fig.~\ref{fig:ablation}, as we progressively apply the schedule primitives to the model, there is a consistent increase in performance.
Kernel optimizations, such as Flash Attention~\cite{dao2022flashattention} and fusing the \texttt{Bias-GeLU} kernel, have shown a 1.18$\times$ speedup on a single device.
By sharding the \texttt{attention} and \texttt{ffn} modules as discussed in Fig.~\ref{fig:schedule_mod}, we can effectively scale out to 8 GPUs, achieving a 4.21$\times$ speedup.
Moreover, since the word embedding layer contains parameters related to a vocabulary size of over 30K, sharding this parameter results in larger batch sizes, leading to a final speedup of 5.69$\times$ and improved scalability when compared to Megatron-LM BERT in Fig.~\ref{fig:8_v100}.}

\begin{figure}[t]
\centering
\includegraphics[width=\linewidth]{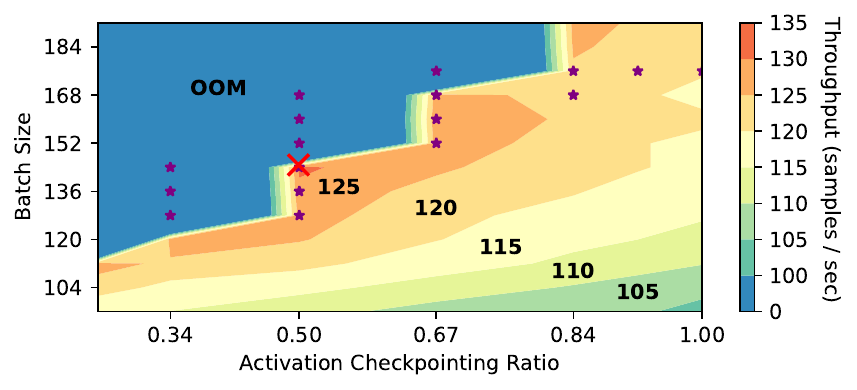}
\caption{Auto-tuning an OPT-350M model. Contour lines show the throughput of different combinations of batch size and checkpoint ratio. Throughput 0 means OOM. Purple \textcolor{purple}{$\star$} indicates the explored configurations via coordinate descent; red \textcolor{red}{$\times$} depicts the optimal one.}
\label{fig:autotune}
\end{figure}

\noindent\textbf{Auto-Tuning.}
\newtext{
To showcase the effectiveness of \Name auto-tuner within a large search space, we assess the performance of an OPT-350M~\cite{zhang2022opt} model using 8 V100 GPUs.
We define a search space (including white and yellow regions in Fig.~\ref{fig:polygon}) with 91 configurations composed of various batch sizes and activation checkpointing ratios.
As shown in Fig.~\ref{fig:autotune}, the optimal configuration only checkpoints 50\% of the layers with the batch size below the memory threshold, and attains over 30\% improved performance compared to the poorest configuration within the search space.
With the search space that has already pruned many inefficient configurations, and the coordinate descent algorithm deployed by the \Name auto-tuner, only 17 configurations (19\% of the total candidates) are explored to identify the optimal configuration with the highest throughput.
The entire search process is completed in 20 minutes, as opposed to the 139 minutes an exhaustive search would require, reducing 86\% of the search time.
}

\section{Related Work}
\label{sec:related_work}

\noindent\textbf{Schedule Languages and Compilers.}
Many domain-specific languages (DSLs) leverage the idea of decoupling optimizations from algorithms~\cite{jonathan2013halide,chen2018tvm,baghdadi2019tiramisu,lai2019heterocl,xiang2022heteroflow,hagedorn2020fireiron,chen2021krill}, allowing users to focus on customization and enabling compilers to perform complex optimizations.
TVM~\cite{chen2018tvm,feng2022tensorir} inherits the decoupling idea from Halide~\cite{jonathan2013halide} and builds an end-to-end compiler for deep learning inference.
\Name borrows a similar decoupling idea and applies it to the model execution level.


\noindent\textbf{Dynamic Graph Optimizations.}
Due to the dynamic nature and usability of PyTorch, many frameworks and libraries directly explore optimizations on top of it.
ZeRO~\cite{rajbhandari2020zero} is a three-stage data parallelism strategy that partitions optimizer states, gradients, and parameters to reduce memory usage, implemented first in DeepSpeed~\cite{rasley2020deepspeed} and then adopted by other frameworks~\cite{zhao2023fsdp}.
MiCS~\cite{zhang2022mics} further improves ZeRO by minimizing the communication scale.
Megatron-LM~\cite{shoeybi2019megatron,narayanan2021megatronv2} takes a different approach to implementing model parallelism for Transformers, becoming one of the mainstream parallelism libraries.
\Name provides comprehensive primitives to apply optimization techniques in a systematic and productive way. 
We also use framework dialects described in \S\ref{sec:impl} to train the scheduled models on these frameworks.
 
\noindent\textbf{Static Graph Optimizations.}
Some other DL frameworks adopt static graphs so that compiler optimizations can be easily involved.
JAX~\cite{jax2018github} is a popular framework that offers a programming model similar to NumPy, and is powered by XLA~\cite{xla} as the backend compiler.
Accordingly, it is able to achieve 3D parallelism with the corresponding sharding mechanism, GSPMD~\cite{xu2021gspmd} and GShard~\cite{lepikhin2021gshard}.
On top of that, Alpa~\cite{zheng2022alpa} is the first compiler based on JAX and XLA to achieve automatic 3D parallelism.
Besides, Unity~\cite{colin2022unity} is another compiler-based distributed training framework that automatically jointly optimizes model parallelism and operator fusion.
Their automation mechanisms are orthogonal to \Name and could inspire \Name's auto-scheduler in the future.

Moreover, PyTorch 2.0~\cite{pytorch20} utilizes \fx~\cite{reed2022fx} as the IR to capture dynamic graphs and perform optimizations like operator fusion.
\newtext{Nevertheless, it lacks native support for 3D parallelism, partial activation checkpointing, etc., which are crucial for training large models.
\Name takes the heavy lifting work from the users, offering a systematic approach for efficient multi-device model optimization and performance improvement through auto-tuning.}
\section{Conclusion}
\label{sec:conclusion}
In this paper, we propose a schedule language \Name for progressive optimization of large model training.
\Name decouples model execution from definition, and provides a comprehensive set of schedule primitives for users to efficiently optimize the model execution.
Experimental results show \Name can combine existing optimizations to align or even outperform their performance with minimal programming effort.
\Name also facilitates the rapid prototyping of emerging optimizations for large model training.
\section*{Acknowledgments}
We would like to thank Jie Wang, anonymous reviewers, and our shepherd, Aditya Kanade, for their insightful feedback.
Hongzheng Chen and Zhiru Zhang are supported in part by ACE, one of the seven centers in JUMP 2.0, a Semiconductor Research Corporation (SRC) program sponsored by DARPA and NSF Award \#2019306.

\newpage
\bibliographystyle{plain}
\bibliography{asplos24}

\newpage
\appendix
\section{Example Schedule on BERT}
\label{appendix:bert}
The following code snippet shows an example schedule on BERT model using \Name, where the 21 lines of schedule code are highlighted.

\bigskip
\begin{lstlisting}[
  language=Python,
  numbers=left,
  backgroundcolor=\color{white},
  breaklines=true,
  linebackgroundcolor={%
    \ifnum\value{lstnumber}<16
      \ifnum\value{lstnumber}>11
        \color{yellow!50}
      \fi
    \fi
    \ifnum\value{lstnumber}<26
      \ifnum\value{lstnumber}>18
        \color{yellow!50}
      \fi
    \fi
    \ifnum\value{lstnumber}<35
      \ifnum\value{lstnumber}>27
        \color{yellow!50}
      \fi
    \fi
    \ifnum\value{lstnumber}<39
      \ifnum\value{lstnumber}>35
        \color{yellow!50}
      \fi
    \fi
    \ifnum\value{lstnumber}>39
      \ifnum\value{lstnumber}<42
        \color{yellow!50}
      \fi
    \fi
    \ifnum\value{lstnumber}<46
      \ifnum\value{lstnumber}>42
        \color{yellow!50}
      \fi
    \fi
    \ifnum\value{lstnumber}<53
      \ifnum\value{lstnumber}>48
        \color{yellow!50}
      \fi
    \fi
  }
]
from transformers import BertLMHeadModel, AutoConfig
config = AutoConfig.from_pretrained("bert-large-uncased")
model = BertLMHeadModel(config)

import slapo
from slapo.pattern import call_module
import torch.nn.functional as F

sch = slapo.create_schedule(model)

# Shard embeddings
sch["embeddings.word_embeddings"].sync(mode="fwd_pre",
    sync_op_or_fn=slapo.op.embed_fwd_hook)
sch["embeddings.word_embeddings"].sync(mode="fwd_post",
    sync_op_or_fn=slapo.op.embed_bwd_hook)
for idx in range(config.num_hidden_layers):
    # Shard self attention module
    subsch = sch[f"encoder.layer.{idx}.attention"]
    subsch["self.query"].shard(["weight", "bias"], axis=0)
    subsch["self.key"].shard(["weight", "bias"], axis=0)
    subsch["self.value"].shard(["weight", "bias"], axis=0)
    subsch.sync(mode="bwd_post", sync_op_or_fn="all_reduce")
    subsch["output.dense"].shard("weight", axis=1)
    subsch["output.dense"].sync("fwd_post",
        sync_op_or_fn="all_reduce")
    # Shard MLP module
    subsch = sch[f"encoder.layer.{idx}"]
    subsch["intermediate.dense"].shard(
        ["weight", "bias"], axis=0)
    subsch["intermediate.dense"].sync("bwd_post",
        sync_op_or_fn="all_reduce")
    subsch["output.dense"].shard("weight", axis=1)
    subsch["output.dense"].sync("fwd_post",
        sync_op_or_fn="all_reduce")
    # Decompose linear bias and trace module
    subsch["attention.output.dense"].decompose()
    subsch["output.dense"].decompose()
    subsch.trace(tracer="huggingface", flatten=True)
    # Replace scaled dot product attention
    subgraphs = subsch.find(slapo.pattern.scaled_dot_product)
    subsch.replace(F.scaled_dot_product_attention, subgraphs)
    # Fuse linear bias and gelu
    subgraph = subsch.find(lambda x, bias: F.gelu(bias + x))
    subsch.fuse(subgraph,
        compiler="TorchInductor", name="BiasGeLU")
    # Fuse bias add, layer norm, and residual
    for ln in ["attention.output.LayerNorm",
                "output.LayerNorm"]:
        subgraph = subsch.find(lambda x, bias, residual:
            call_module(ln, F.dropout(bias + x) + residual))
        subsch.fuse(subgraph,
            compiler="TorchInductor", name="LNResidual")
\end{lstlisting}

\end{document}